\documentclass[preprint,12pt]{elsarticle}

\usepackage{amssymb}

\usepackage[utf8]{inputenc}
\usepackage{amsmath,amsfonts}
\usepackage{graphicx}
\usepackage{textcomp}

\usepackage{multirow}
\usepackage{times}
\usepackage{soul}
\usepackage{url}
\usepackage[hidelinks]{hyperref}
\usepackage[utf8]{inputenc}
\usepackage[small]{caption}
\usepackage{amsthm}
\usepackage{booktabs}
\usepackage{algorithm}
\usepackage{amssymb}
\usepackage[table]{xcolor}
\usepackage{xcolor}
\usepackage{svg}
\usepackage[normalem]{ulem}
\usepackage{algpseudocode}

\journal{Neural Networks}

\begin{document}

\begin{frontmatter}


\title{Deep Neural Networks as Complex Networks}

\author[inst1]{\footnote{Equal contribution.}Emanuele La Malfa}
\author[inst2]{$^{1}$Gabriele La Malfa}
\author[inst3]{Claudio Caprioli}
\author[inst3]{\footnote{Corresponding author. For inquiries, please write at emanuele.lamalfa@cs.ox.ac.uk.}Giuseppe Nicosia}
\author[inst3,inst4]{$^{2}$Vito Latora}

\affiliation[inst1]{organization={Department of Computer Science},
            addressline=University of Oxford}
\affiliation[inst2]{organization={Department of Informatics},
            addressline=King's College London}
\affiliation[inst3]{organization={The Department of Biomedical \& Biotechnological Sciences},
            addressline=University of Catania}
\affiliation[inst4]{organization={The Department of Physics and Astronomy "Ettore Majorana"},
            addressline=University of Catania}
\affiliation[inst5]{organization={School of Mathematical Sciences},
            addressline=Queen Mary University of London}


\begin{abstract}
Deep Neural Networks are, from a physical perspective, graphs whose `links` and `vertices` iteratively process data and solve tasks sub-optimally. We use Complex Network Theory (CNT) to represents Deep Neural Networks (DNNs) as directed weighted graphs: within this framework, we introduce metrics to study DNNs as dynamical systems, with a granularity that spans from weights to layers, including neurons. CNT discriminates networks that differ in the number of parameters and neurons, the type of hidden layers and activations, and the objective task. We further show that our metrics discriminate low vs. high performing networks. CNT is a comprehensive method to reason about DNNs and a complementary approach to explain a model's behavior that is physically grounded to networks theory and goes beyond the well-studied input-output relation.
\end{abstract}

\begin{keyword}
Neural Networks \sep Complex Networks \sep Complex Networks Theory
\end{keyword}

\end{frontmatter}


\section{Introduction}
Deep Neural Networks (DNNs) have contributed in the recent years to the most remarkable progress in artificial intelligence. Algorithms now reach human-comparable (or even super-human) performances in relevant tasks such as computer vision, Natural Language Processing etc.~\cite{GOODFELLOW2016}. Nonetheless, it is still unclear how neural networks encode knowledge about a specific task: interpretability has thus become increasingly popular~\cite{MONTAVON2018} as the key to understanding enabling factors behind DNNs remarkable performances. 
\newline 
Complex Network Theory~\cite{BOCCALETTI2006} (CNT) is a branch of mathematics that investigates complex systems, from the human brain to networks of computers, by modeling and then simulating their dynamics through graphs where nodes are entities and vertices relationships~\cite{CHAVEZ2010, CRUCITTI2004, PORTA2006}.
\newline \newline
In this work, we address the problem of characterizing a Deep Neural Network as a graph with ad-hoc CNT metrics. We formalize the approach by identifying a corpus of metrics that describe peculiarly weights', neurons' and hidden layers of a network. By comparing `snapshots` of such metrics from a network -- or a population of networks -- we spot trends that generalize across different architectures, initialization and objective tasks. CNT metrics can represent architecturally different topologies, as we illustrate for Fully Connected (FC), Auto-Encoders (AE), Convolutional (CNNs) \cite{LECUN1995} and Recurrent Neural Networks (RNN) \cite{HOCHREITER1997}. We conduct an extensive experimental evaluation to populations of FC and AE networks where we vary the architectural details -- i.e., hidden layers, number of neurons and activation functions -- on image classification and signal reconstruction tasks. Finally, we release a fast and extensible package to replicate the experiments and further extend CNT metrics beyond the application of this paper.\footnote{Code will be released soon.}
\newline \newline
Concisely, CNT neural network's dynamics are capable of: \textbf{(i)} identifying specific task-dependent patterns in MNIST and CIFAR10, in case of classification as well as image reconstruction; \textbf{(ii)} Discriminating the between different DNNs activation functions (i.e., linear, ReLU and sigmoid) of shallow and deep architectures; \textbf{(iii)} Revealing the performance gaps of trained vs. untrained networks.
\newline
We believe the application of the CNT framework presented in this paper can stimulate further research in general-purpose Deep Learning, e.g., enhancing our understanding of different architectures such as CNNs, RNNs, attention, etc. 
\newline

\section{Related Works}
We structure this Section with the following rationale: we first discuss in details those works where Complex Networks Metrics are defined, applied and/or extended to study neural networks' dynamics, precisely mentioning where our work differs. We then focus on those research papers where CNT is used to enhance a neural networks, e.g., by extract input features or `bootstrapping` its training parameters.
\newline
A few recent works have addressed the problem of analysing neural networks as directed graphs within the Complex Network Theory. In \cite{LAMALFA2021}, the authors propose an evaluation of the training dynamics of Deep Neural Networks via Complex Network metrics. Their approach encompasses populations and single instance networks: they define metrics at different level of granularity that they evaluate on computer vision tasks. Differently from our work, their metrics assume the value of the input as static, thus making it hard to judge results that come from different initial settings (i..e, different architectures, initializations, number of hidden layers etc.). Furthermore, they propose metrics that do not directly depend on the input data but solely on the state of the network at a precise training step, while we fill this gap with 2 ad-hoc metrics. 
\newline
An application of CNT metrics to feed-forward neural networks is described in~\cite{ZAMBRA2020}, where the authors focus on the emergence of connections between neurons that present both an interesting geometric shape and strong values of the corresponding Link-Weights. Differently from our work, in~\cite{ZAMBRA2020} the authors consider advanced CNT metrics such as \textit{motifs} (e.g., triangles between link-weights of adjacent layers), while our intent is firstly to settle the vocabulary of basic correspondences -- i.e., link/weights, nodes/neurons, layers -- between Deep Neural Networks as dynamic systems and Complex Network Theory.
\newline
In \cite{PETRI2021}, a formal analysis is conducted to assess the parallel processing capability of neural architectures via CNT. With respect to our work, this one develops a parallelism between Deep Networks and CNT that concerns mainly the number of tasks that can be learnt concurrently, while in our work we train multiple models on a single-task at a time with the intent to discriminate different networks optimized on the same problem.
\newline
The authors of~\cite{TESTOLIN2019} apples CNT to distill information from Deep Belief Networks: Deep Belief Networks - which are generative models that differ from feed-forward neural networks as the learning phase is generally unsupervised - are studied with the lens of CNT by turning their architectures into a stack of equivalent Restricted Boltzmann Machines. An irreconcilable difference with our approach is the employment of Restricted Boltzmann Machines, which despite constitutes a milestone in the advancement of modern Deep Learning, they have seen their interest by the research community decline in the recent years in favor of other architectures such as FCs, CNNs, etc.
\newline
In \cite{SCABINI2021}, the authors propose Complex Network techniques to analyze the structure and performance of fully connected neural networks, showing high correlation to the networks classification performances. While their research questions partially overlaps with ours, the authors develop a theory and a framework to identify similar neurons that is based on centrality measures and encompasses FC networks. Our approach consider metrics at weights, neurons and layers level and despite involves comparisons between high and low-performing networks, it develops across further discriminant factors such as depth, hidden activations, architecture and task.
\newline
In~\cite{SAXE2022}, the authors derive a framework that schematizes how the information flow impacts learning dynamics, in a few cased identifying exact dynamics in a process that can be generalized and that they denote as `neural race`. While our contribution is partially focused on neural networks' learning dynamics, the scope of our work is to provide a systematic background, i.e., a `syntax` and a `vocabulary`, to reason of DNNs via complex networks.
\newline 
To complete the picture, a number of works have applied CNT with the objective to enhance a network's training phase: examples are \cite{SCABINI2022,RIBAS2020}.\newline

\section{Methodology} \label{sec:ctn-metrics}
This Section provides a concise framework to study DNNs via CNT. We begin by presenting the notation that will be used throughout the paper: a subsection is dedicated to describing the Fully Connected topology (FC), which will serve as a basis to  represent Convolutional and Recurrent topologies in the CNT framework. 
\newline 
We then introduce, define and describe several CNT metrics: we analytically derive the metrics distributions that we expect from untrained DNNs whose weights are initialized according to a known probability distribution. As CNT metrics apply to any neural network architecture that can be represented as a directed graph, we conclude the Section with a formal approach to efficiently represent and analyse Convolution and Recurrent layers via CNT. 

\subsection{Deep Neural Networks Background}
\begin{figure}
    \centering
    \includegraphics[width=1\linewidth]{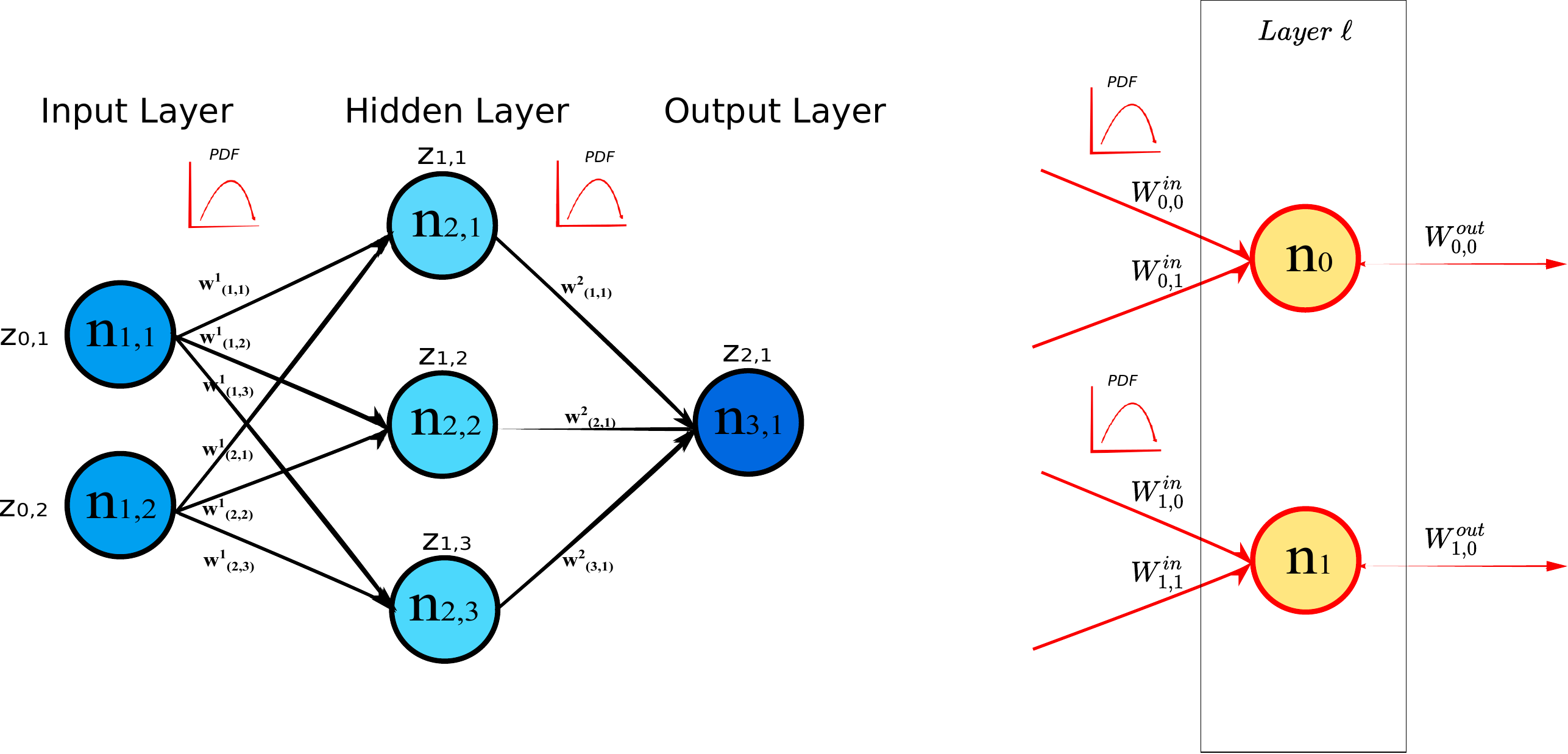}
    \caption{An FC network is represented as a Complex Network. Links have weights assigned to them, while neurons are the vertices of the graph where the `computation` happens (left). Neurons are stacked hierarchically into layers (right).}
    \label{fig:fcgraph}
\end{figure}

We consider a Fully Connected network (FC) that solves a supervised task, i.e., the network learns an input-output mapping $f: \mathbb{R}^d \xrightarrow{} \mathbb{R}^m$ that minimizes a generic loss function $\mathcal{L}(f(x), y)$, with $(x,y)$ each pair of input-output. An input $x$ is a $d$-dimensional vector $x \in \mathbb{R}^d$ drawn from a distribution, while each corresponding output is either from a discrete set in case of classification, i.e., $c \in C \ . \ |C|=m$, or it is continuous in case of regression, i.e., $y \in \mathbb{R}^m$.
\newline
An FC architecture consists of $L>0$ dense layers stacked together, each of a variable number of neurons: within each hidden layer $\ell$, a neuron $n^{[\ell]}_{i}$ is connected through a weighted link to all the neurons of the successive layer $\ell + 1$.
The output $z^{[\ell]}$ of a layer $\ell$ is the product of an affine transformation between a matrix of weights $\Omega^{[\ell]}$, plus eventually a bias term $\beta^{[\ell]}$, namely $z^{[\ell]} \ = \ z^{[\ell-1]}\Omega^{[\ell]} \ + \ \beta^{[\ell]}$, followed by a non-linear activation function $f^{[\ell]}(z^{[\ell]})$.
For an FC network, $\Omega^{[\ell]}$ is a matrix of size $\mathcal{N}^{[\ell]}\times\mathcal{N}^{[\ell+1]}$ and $\beta^{[\ell]}$ is a vector of size $\mathcal{N}^{[\ell+1]}$.
\newline 
To compact the notation, the input and output vectors can be referred as $x=z_0$ and $y = z^{[L]}$, while $\Omega^{[\ell]}$ and $\beta^{[\ell]}$ in the previous formulae refer to the parameters of a neural network layer $\ell$. In this sense, the output of the neural network is hence defined as $y=f^{[L]}(z^{[L]}) = z^{[L-1]}\Omega^{[L]} \ + \ \beta^{[L]}$. We assume, without loss of generality, that for classification $y$, or equivalently $z^{[L]}$, is a vector of real numbers $y \in \mathbb{R}^m$, from which the $argmax$ operator extracts the predicted class, while in case of regression, we just consider the value of each element in $f^{[L]}(z^{[L]})$. Throughout this work, we will also refer to the input-output relation of a neural network at layer $\ell$ as $z^{[\ell]} = \mathbf{f}(x, \ \Omega^{[:\ell]}, \ \beta^{[:\ell]})$.
\newline
As one can formally define the operations computed by each hidden layer $\ell$, in the same way one can analyse the output of each i-th neuron at layer $\ell$, namely $z^{[\ell]}_{i} = f^{[\ell]}(z^{[\ell-1]}\Omega^{[\ell]}_{i} + \beta^{[\ell]}_{i})$, where $\Omega^{[\ell]}_{i}$ identifies the i-th row of $\Omega^{[\ell]}$, while $\beta^{[\ell]}_{i}$ is the i-th element of the bias vector $\beta^{[\ell]}$.
We visualize the information exposed in this paragraph in Figure \ref{fig:fcgraph}.

\subsection{Complex Networks Metrics for Deep Neural Networks}
Within the CNT framework, a neural network is formulated and represented as a directed bipartite graph $Net_{f}(N,E)$. Each vertex $n^{[\ell]}_{i} \ \in \ N, i \in \{0, \mathcal{N}^{[\ell]}\}$ is a neuron that belongs to a hidden layer $\ell$. The intensity of a connection - denoted as "weight" in both CNT and DNNs - is a real number assigned to an edge $(e_{n^{[\ell]}_{i}, n^{[\ell+1]}_{j}} \ \in \ E)$ that connects two neurons. 
Each weight $\omega^{[\ell]}_{i,j}$ is assigned to the link that connects neuron $i$ from layer $\ell$ to neuron $j$ from layer $\ell+1$. 
\newline \newline
On this parallelism we now define metrics that can describe a network as a Complex Network, via a `vocabulary` of CNT metrics that describe a network at different levels of granularity. We report a synopsis of each metric, which precedes an in-depth analysis. 

\subsubsection{Link-Weights Dynamics}
These metrics reflect the dynamic of the parameters in each layer as the training phase evolves toward an optimum. As the network performs the training phase, we investigate weight dynamics in terms of mean and variance in each layer.
Given a neural network layer $\ell$, we define:

\begin{equation}
\mu^{[\ell]} = \dfrac{1}{N^{[\ell]} N^{[\ell+1]} }\sum_{i=1}^{N^{[\ell]}} 
\sum_{j=1}^{N^{[\ell+1]}} \omega^{[\ell]}_{i,j} + \beta^{[\ell]}_i
\end{equation}
\begin{equation}
\delta^{[\ell]} = \dfrac{1}{N^{[\ell]} N^{[\ell+1]} }\sum_{i=1}^{N^{[\ell]}} 
\sum_{j=1}^{N^{[\ell+1]}} ((\omega^{[\ell]}_{i,j} + \beta^{[\ell]}_i ) -\mu^{[\ell]})^2
\end{equation}

The evolution of mean and variance through the training steps gives significant background about learning effectiveness and stability. An underrated yet common issue is when the weights norm does not grow, as it is often a symptom of model over-regularization. On the other hand, it is well known that where the weights grow too much, one may incur in over-fitting problems that are mitigated by regularization techniques.

\subsubsection{Nodes Strength}
The strength $s^{[\ell]}_{k}$ of a neuron $n^{[\ell]}_{k}$ is the sum of the weights of the edges incident in $n^{[\ell]}_{k}$. neural networks graphs are directed, hence there here are two components that contribute to the Node Strength: the sum of the weights of outgoing edges $s^{[\ell]}_{out,k}$, and the sum of the weights of in-going links $s^{[\ell]}_{in,k}$.

\begin{equation} \label{eq:nodes-strength}
s^{[\ell]}_{k} = s^{[\ell]}_{in,k} + s^{[\ell]}_{out,k} =\ \sum_{i=1}^{N^{[\ell]}}(\omega^{[\ell]}_{i,k} + \beta^{[\ell]}_k) + \sum_{j=1}^{N^{[\ell+1]}}\omega^{[\ell+1]}_{k,j}
\end{equation}

Despite analysing the nodes strength as in eq. (\ref{eq:nodes-strength}) is pretty common in Complex Networks, one can consider the input and output strengths as separated, namely $s^{\ell}_{in,k}$ and $s^{\ell}_{out,k}$, and build more neural-specific metrics on top of them. 
\newline 
In the the next subsections, we discuss metrics that depend both on the networks topology and the dataset on which a network has been trained, whose effect we estimate via sampling on form the input data distribution. 

\subsubsection{Neurons Strength}
While the Nodes Strength is a `static` measure of the intensity of a connection, regardless of the value of the inputs, the Neurons Strength encompasses this information via sampling from the input distribution $\mathcal{X}$, or possibly any input distribution that we want to test.
\newline \newline 
\begin{equation}
\zeta^{[\ell]}_{k} =\ \sum_{i=1}^{N^{[\ell]}}z^{[\ell-1]}_{i}\omega^{[\ell]}_{i,k} + \beta^{[\ell]}_{k}, \ \ z^{[\ell-1]} = \mathbf{f}(x, \Omega^{[:\ell]}, \beta^{[:\ell]}) \ . \ x \sim \mathcal{X}
\end{equation}
From a mathematical perspective, the Neurons Strength is a more general version of the Nodes Strength where both the effects of the activation functions -- excluding the layer one is interested in studying -- and the input are considered. 

\subsubsection{Neurons Activation}
Each neurons of a DNN has a value of Neuron Activation that depends on the value of the input and the activation functions, namely
\newline \newline 
\begin{equation}
a^{[\ell]}_{k} =\ f^{[\ell]}(\sum_{i=1}^{N^{[\ell]}}z^{[\ell-1]}_{i}\omega^{[\ell]}_{i,k} + \beta^{[\ell]}_{k}), \ \ z^{[\ell-1]} = \mathbf{f}(x, \Omega^{[:\ell]}, \beta^{[:\ell]}) \ . \ x \sim \mathcal{X}
\end{equation}

When a neuron has an anomalously high value of Node Strength, it is propagating, compared to the other neurons in the layer, a stronger \textit{signal}: this is a hint  that either the network is not able to propagate the signal through all the neurons uniformly or that the neuron is in charge of transmitting a relevant portion of the information for the task. On the other hands, a node that propagates a weak or null \textit{signal} may be pruned (hence reducing the network complexity) as it doesn't contribute significantly to the output of the layer. 

\subsubsection{Layers Fluctuation}
These metrics extend to DNNs the notion of Nodes Fluctuation so that it is possible to measure the Neurons/Nodes metrics at the level of the network hidden layers.
CNT identifies neural network asymmetries at nodes and links level. The standard measure, known as Nodes Disparity \cite{BOCCALETTI2006}, is defined for a node $n^{[\ell]}_i$ as $Y^{[\ell]} = \sum_{i=1}^{N^{[\ell]}}[{\omega^{[\ell]}_{i}}/{s^{[\ell]}_{i}}]^2$. Nodes Disparity ranges from $0.$ to $1.$ with the maximum value when all the weights enter a single link. Conversely, weights that are evenly distributed cause the nodes in the networks to have the same - minimum - value of Disparity.
Nonetheless Disparity and similar metrics are widely adopted for studying Complex Networks, a fundamental problem arises when weights in the previous equation assume positive and negative values, as in the case of DNNs: the denominator can be zero for either very small values of weights or concurrently as the sum of negative and positive values that are equally balanced. In addition, it is appropriate to include a metric that measures the fluctuation of strengths in each layer, as the nodes in a DNN contribute in synergy to the identification of increasingly complex patterns. 
We thus propose a metric to measure the Strength fluctuations in each layer, as a proxy of the complex interactions among nodes at the same depth. \newline
\indent The \textit{Layers Fluctuation} for a DNN at layer $\ell$ is defined as:
\begin{equation}
Y^{[\ell]} = \sqrt{\dfrac{\sum_{i=1}^{N^{[\ell]}}(s^{[\ell]}_{i} - \hat{s}^{[\ell]})^2}{I}}
\end{equation}
 
where $\hat{s}^{[\ell]}$ is computed as the average value of Nodes Strength at layer $\ell$, namely $\hat{s}^{[\ell]}=\dfrac{1}{N^{[\ell]}}\sum^{m}_{i=1}s^{[\ell]}_{i}$, being $N^{[\ell]}$ the number of nodes/neurons at layer $\ell$. 
\newline 
Please note that differently from the standard \textit{Nodes Fluctuation}, the \textit{Layers Fluctuation} formula drops the dependence from each specific node $n_i$ to characterize a layer's dynamics.
The advantage of this metric is to measure disparity in a way that avoids numerical problems yet allowing to describe the networks whose weights can assume any range of values, without being restricted to positives only. \newline
\textit{Layers Fluctuation} can be used to spot bottlenecks in a network, i.e., cases where a layer impedes the information from flowing uninterrupted through the architecture. In the experimental evaluation we show how Layers Fluctuation enables to spot interesting behaviors of a network while other metrics (included the Link-Weights and Nodes Strength) are not sufficient.

\subsection{Adapting CNT Metrics Beyond the Fully Connected Topology}\label{sec:cnn-metrics}
\begin{figure}
    \centering
    \includegraphics[width=1\linewidth]{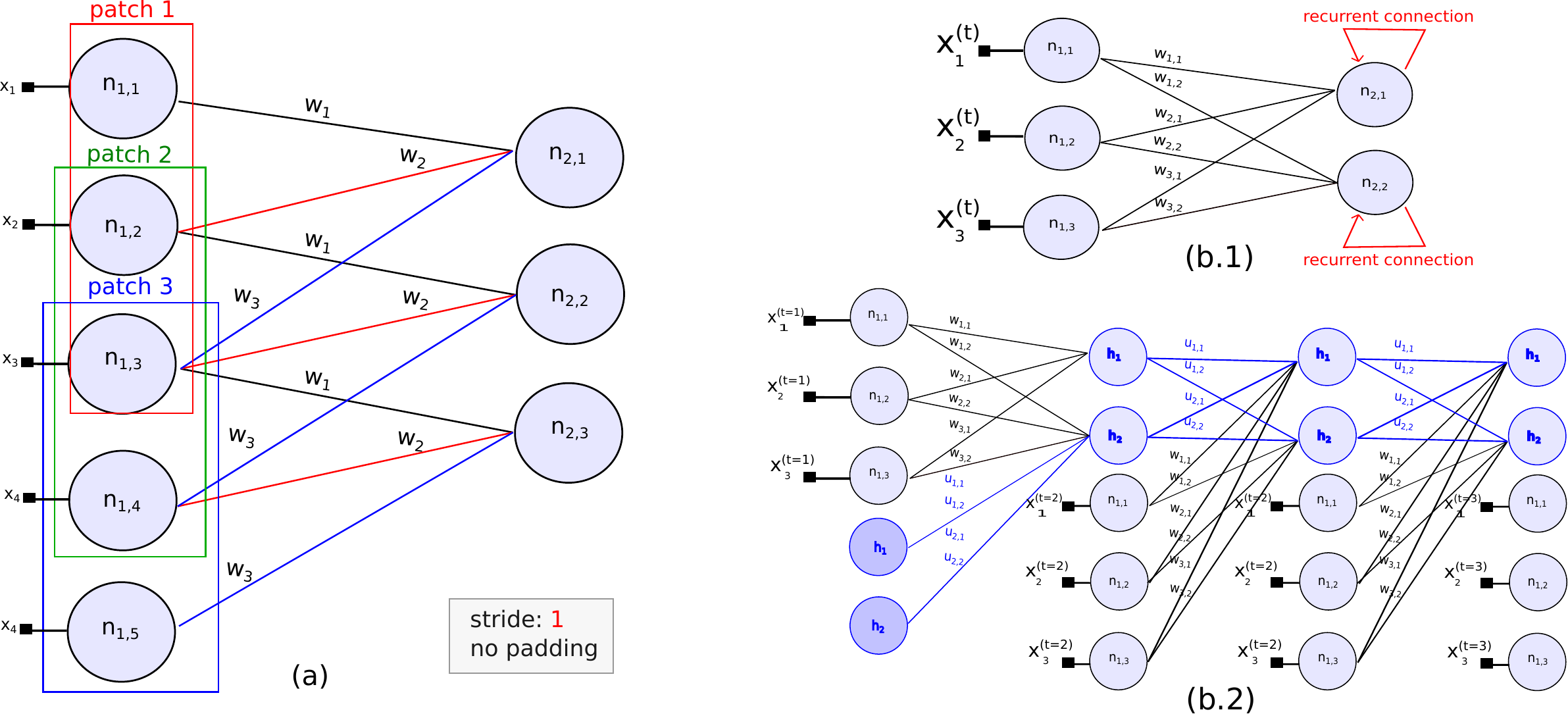}
    \caption{On subplot (a), a representation of how convolution is represented via CNT: please notice the shared kernel's weights which contribute independently to the convolution with each input's sub-patch. On subplot (b.1) a recurrent layer that is turned into a recurrent layers via network unfolding and can then be analysed via CNT (subplot (b.2)).}
    \label{fig:cnn-rnn-cnt}
\end{figure}

The evidence that architectural inductive biases helped improving performances of DL models is overwhelming, as testified by decades of research in these directions. Just to name a few works that have impacted the entire field of artificial intelligence, CNNs have been built with biases to local connectivity to mimic the human's vision system \cite{LECUN1995}, while recurrent networks formalize gates and memory cells to keep the information readily available at distant time-steps~\cite{hochreiter1997long}. 
\newline 
In this Section we show how to adapt the CNT metrics proposed in this paper beyond the FC topology. We select Convolution and recurrent neural networks, as they constitute the building blocks of thousands of specialized architectures that allowed DNNs to reach and surpass human capabilities in many tasks in in vision and language processing.
\newline \newline 
\noindent \textbf{Convolution. } We consider the operation of convolution between an input matrix $z_{\ell}$ of $w*h$ numbers and a kernel of size $k*k, \ s.t. \ k<w \wedge k<h$, namely $z_{\ell+1}=conv(z_{\ell}, k)$. Convolution is optionally followed by an activation function. In order to adapt CNT metrics to this operation, a straightforward strategy is to turn convolution into an equivalent dot product between a vector and a matrix via the Toeplitz matrix. Despite its simplicity, this approach comes with the cons of a quadratic increase in the complexity of the algorithm and it is thus infeasible for large networks. We instead approach the problem by first identifying and isolating each portion of the input that is multiplied, independently and via an element-wise product, to the convolutional kernel: by coupling each input neurons (see Figure \ref{fig:cnn-rnn-cnt}, subplot (a)) with the respective output neuron, we can calculate the metrics for any layer with a much faster algorithm.
\newline 

\noindent \textbf{Recurrent Cell. } Recurrent neural networks allow to process an input sequentially, so that the output of each sequence depends, recursively, to the output of the the previous one. We propose to distill the CNT metrics for RNNs by unfolding each recurrent unit (see Figure \ref{fig:cnn-rnn-cnt}, subplots (b.1) and (b.2)) and thus reducing the problem to the simpler case of an FC network. 

\subsection{Exact Metrics of Untrained Networks} \label{sec:theory}
From a theoretical perspective, one can derive the exact distributions of the CNT metrics of a network before it is trained on a task. By assuming parameters $\Omega,\beta$ are drawn from a know distribution. In this Section we show how  the choice of the distribution from which parameters $\Omega, \beta$ are drawn model the Nodes Strength and the Fluctuation.\footnote{In principle, one can also assume datapoints $x \in \mathcal{X}$ are drawn from a known distribution and perform the analysis for Neurons Strength and Activation. In practice, we show that even the exact computation of the initial distributions of Strength and Fluctuations is not trivial.}
\newline \newline
We consider a $L$-layers FC network whose weights are sampled independently from a Gaussian distribution with finite variance, namely $w^{[\ell]}_{i,j} \sim \mathbf{N}(0, \sigma^2)$, while biases are, for sake of simplicity, set to zero. We show how to derive the CNT metrics of a generic hidden layer $\ell$ network where the weights of two adjacent layers are initialized sampling parameters from the same known distribution. It emerges clear from this analysis how hard is to treat CNT metrics for DNNs as statistical objects, thus the necessity of an approach that is empirical, despite rooted on math.
\newline \newline 

\noindent \textbf{Link-weights.} As $w^{[\ell]}_{i,j} \sim \mathbf{N}(0, \sigma^2)$, the PDF of the Link-weights that we expect to see is a gaussian centered in zero and with variance $\sigma^2$. Same for $w^{[\ell]}_{i,j}$.
\newline 

\noindent \textbf{Nodes Strength.} We derive the Nodes Strength of a neuron n as the sum of the input and out strengths, namely $s^{[\ell]}_i=s^{[\ell]}_{in,i}+s^{[\ell]}_{out,i}$
. As each weight is sampled independently from the others, $s^{[\ell]}_i \sim N(0, K\sigma^2)$, with $K=I+J$, where $I$ and $J$ are respectively the in/out degrees of neuron $n^{[\ell]}_i$ (i.e., the number of incoming and out-coming links).
\newline
Interestingly, the average value of Node Strength from a layer $\ell$ will have the variance that grows linearly with the number of incoming and out-coming links to each node, thus resulting in a metric that is independent from the number of neurons of that hidden layer: in fact its variance is directly dependent on the number of neurons in the precedent and successive layers. With this initialization, one can indeed identify `bottlenecks` in a DNN by just analysing the distributions of the Node Strengths at adjacent hidden layers.
\newline

\noindent \textbf{Nodes Fluctuation.} We conclude the Section showing how to compute the fluctuation of the input strengths, assuming there are K=I+J neurons at layer $\ell$.\footnote{For the sake of clarity, in this subsection we will denote the number of neurons at layers $\ell$ and $\ell$+1 -- referred elsewhere as $N^{[\ell]}$ and $N^{[\ell+1]}$ -- with $I$ and $J$, not to confuse the reader with the notation used for the Normal distribution $\mathbf{N}$.} Being $Y^{[\ell]}$ the the Nodes Fluctuation at layer $\ell$, and denoting the set of nodes at layer $\ell$ as $S^{[\ell]}=\{s^{[\ell]}_1, .., s^{[\ell]}_J\}$), we expand $Y^{[\ell]}$ so that $Y^{[\ell]}=\sqrt{\dfrac{\sum^{K}_{k}(s^{[\ell]}_k - \hat{s^{[\ell]}})^2}{K}}$, where $\hat{s^{[\ell]}}=\dfrac{\sum^{K}_{k}s^{[\ell]}_k}{K}$. Now $\hat{s^{[\ell]}} \sim \mathbf{N}(0, \sigma^2)$, hence 
$(s^{[\ell]}_k - \hat{s^{[\ell]}}) \sim \mathbf{N}(0, \sigma^2(K-1))$.
We now call $s'^{[\ell]}_k = (s^{[\ell]}_k - \hat{s^{[\ell]}}) \sim \mathbf{N}(0, \sigma^2(K-1))$.
\newline Let $q^{[\ell]}_k = \dfrac{s^{[\ell]}_k}{\sqrt{K}\sigma}$ be the re-scaled strength, such that $q^{[\ell]}_{k} \sim \mathbf{N}(0, 1)$. We can rewrite $(s^{\ell}_k - \hat{s^{[\ell]}})$ as 
$\dfrac{\sqrt{K}\sigma}{\sqrt{K}\sigma}(s^{[\ell]}_k - \hat{s^{[\ell]}}) = \sqrt{K}\sigma (q^{[\ell]}_k - \hat{q^{[\ell]}})$, where $\hat{q^{[\ell]}} = \sum^{K}_{k}\dfrac{1}{K} \dfrac{s^{[\ell]}_k}{\sqrt{K}\sigma}$.\newline 
Let's consider now the squared fluctuation $(Y^{[\ell]})^2= \sum^{K}_{k}{\dfrac{(s^{[\ell]}_k - \hat{s^{[\ell]}})^2}{K}}$. We can rewrite it as a function of the difference $(q^{[\ell]}_k - \hat{q^{[\ell]}})$ as: 

\begin{equation}
    (Y^{[\ell]})^2=\sigma^2 \sum^{K}_{k}(q^{[\ell]}_k - \hat{q^{[\ell]}})^2
\end{equation}

According to Cochran's theorem the sum $\sum^{K}_{k}(q^{[\ell]}_k - \hat{q^{[\ell]}})^2$  follows a Chi-squared $\chi$ distribution with $K-1$ degrees of freedom. We conclude that: 

\begin{itemize}
    \item $(Y^{[\ell]})^2$ is the product of a $\chi^2(K-1)$ and a constant, but it's not a Chi-squared random variable.
    \item $\dfrac{(Y^{[\ell]})^2}{\sigma^2} \sim \chi^2(K-1)$, i.e., a Chi-squared distribution with $K-1$ degrees of freedom.
\end{itemize}

In summary, we believe that an empirical analysis of DNNs metrics is cogent for the following reasons: \textbf{(i)} it is hard to estimate what is the distribution of the parameters of a CNT metric, if any; \textbf{(ii)} During the training phase, parameters change and correlate, thus nullifying the i.i.d. hypothesis on the initial distribution of parameters.

\section{Experimental Evaluation}
We conduct the experimental evaluation according to the following precise pipeline: \textbf{(i)} we choose an architecture (FC, AE), its depth in terms of hidden layers (3, 7), and a set of activation functions (linear, ReLU, sigmoid); \textbf{(ii)} We initialize a pool of $n=30$ neural networks, whose parameters we bootstrap from a Gaussian distribution of known variance ($0.5$ from MNIST, $0.05$ for CIFAR10), that we then train on a task (MNIST, CIFAR10): on these datasets, for FC architectures we perform image classification, while AEs reconstruct the original input/signal; \textbf{(iii)} We compute, aggregate and plot the CNT metrics for each population of networks.
\newline
As CIFAR10 is difficult to solve with high accuracy with solely FC architectures~\cite{LIN2015}, we further compare how the training phase varies between high vs. low-performing networks. 
\footnote{We note that the same approach could be applied to MNIST, however it is difficult to obtain 'under-performing' neural networks, as even simple architectures reach levels of accuracy well beyond $90\%$. The reader can refer to~\cite{LAMALFA2021} for an in-depth discussion of the topic.}
\newline
Below we provide a bulleted list of the progression of the Section to assist in interpreting and comparing the results. Each experimental evaluation answers a specific research question (RQ) that we formulate alongside.
\begin{enumerate}
        \item CNT metrics for task discrimination.
        \newline
        \textbf{RQ1.} How do CNT metrics differentiate MNIST from CIFAR10 and, more generically, two learning tasks?
        \item Sensitivity of CNT metrics to neural networks' activations.
        \newline 
        \textbf{RQ2.} How do CNT metrics respond to different activation functions?
        \item Sensitivity of CNT metrics to neural networks' depth.
        \newline 
        \textbf{RQ3.} Are CNT metrics effective at describing the behaviour of neural networks with different depth (number of layers)?
        \item CNT metrics of low vs. high-performing networks.
        \newline 
        \textbf{RQ4.} Do CNT metrics discriminate networks that solve a task with different levels of accuracy?
\end{enumerate}

In the next Sections, we present some empirical evidence that CNT can answer each research question. We also point out that many more results are available as part of the supplementary material in the code, as a full report would have made the Section unnecessarily prolonged.  

\subsection{Results on Classification}
\begin{figure}[]
    \centering
    \makebox[\textwidth][c]{\includegraphics[width=1.2\textwidth]{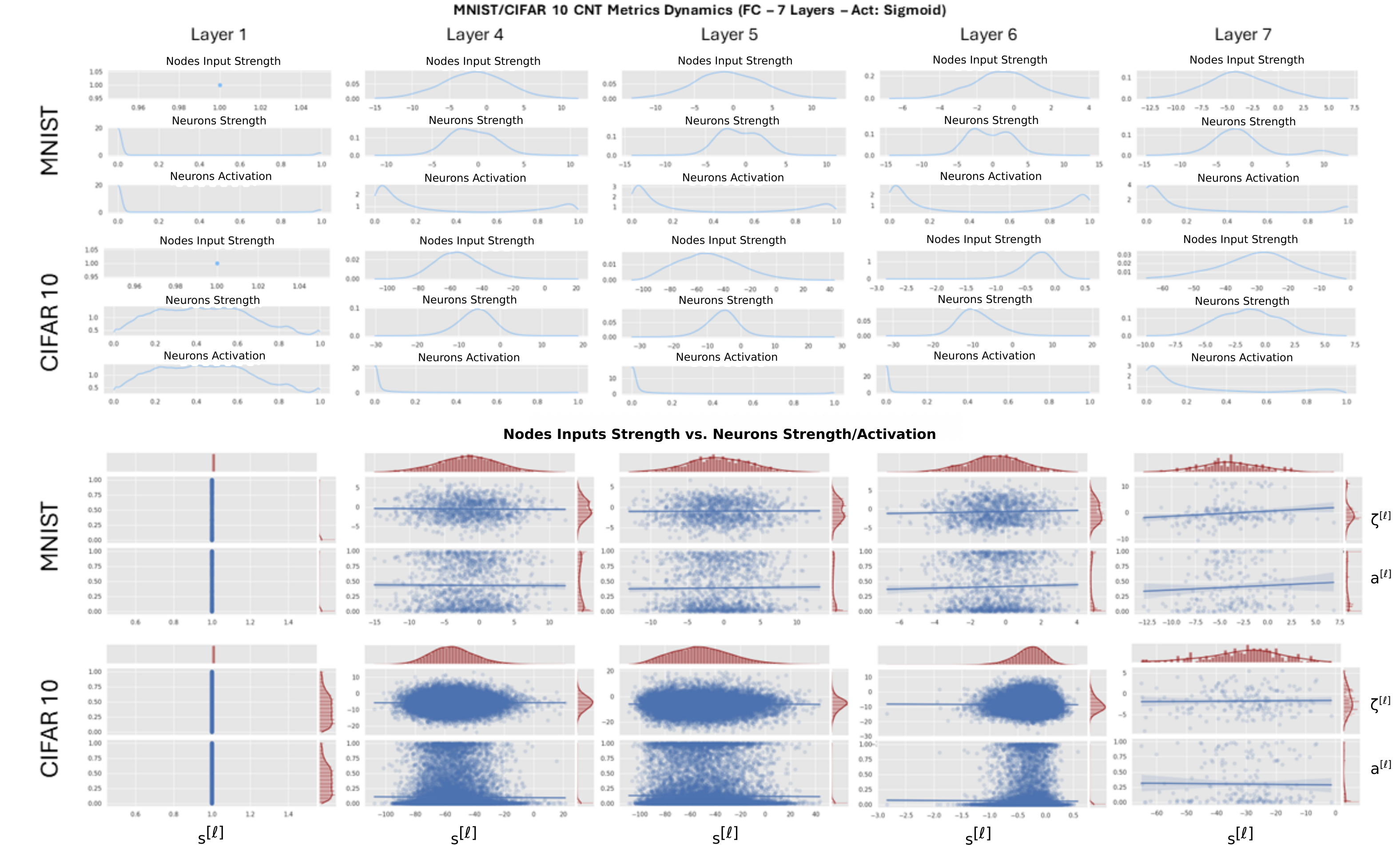}}%
    \caption{CNT metrics to discriminate MNIST from CIFAR10 (activation: sigmoid, 7 hidden layers).}
    \label{fig2}
\end{figure}

\paragraph{\textbf{Tasks discrimination}}
In Fig.\ref{fig2}, we report how CNT discriminate between the two tasks, namely MNIST and CIFAR10.  
As prescribed in the Methodology, networks with fewer neurons (i.e., MNIST) have Nodes Input Strength with reduced support, compared to networks trained on CIFAR10 where it is significantly larger. Interestingly, the bell-shape of the Nodes Strength is preserved in both the tasks. While for MNIST the support of the Nodes Strength is similar among all the layers, at layer $5$ of CIFAR10 we notice a significant reduction of such support: this phenomenon is a bottleneck which possibly impedes the network from performing well.  
\newline
In Neurons Strength, there is a clear trend toward bimodality from layer $4$ for MNIST with a sigmoid activation while this in not true in CIFAR10. 
However, the same observation is not valid with ReLU and linear activations, probably meaning that sigmoid has an intrinsic bias towards bimodality. 
\newline
While Nodes Strength is a singleton in any first layer of a network, both Neurons Activation and Neurons Strength show how different are MNIST and CIFAR10 in the distributions of the inputs. In fact, MNIST first layer shows a bi-modal distribution around $0$ (black input pixels) and $1$ (white input pixels), with greater density around $0$. On the other hand, CIFAR10 distribution shows many more modalities over its support with a smaller peak around $1$, a hint that the task is more complex.
\newline
As regards the deepest layers, the Neurons Activation shows a similar trend for both MNIST and CIFAR10 with the difference that MNIST always shows a marked bimodality around $0$ and $1$. For CIFAR10 the Neurons Strength flattens out with mass concentrated around $0$ and a tail to the right.
\newline
The plots on bottom report scatter plots of correlations between Nodes Strength and, respectively, Neurons Strength and Neurons Activation.
There is weak/no-correlation between the Nodes and Neurons Strength,
which suggests that the input induces independency between the two metrics. On the other hands, we notice a concentration of Nodes Strength and Neurons Activation around $0$ and $1$, i.e., the saturation points of the sigmoid.

\begin{figure}[]
    \centering
    \makebox[\textwidth][c]{\includegraphics[width=1.2\textwidth]{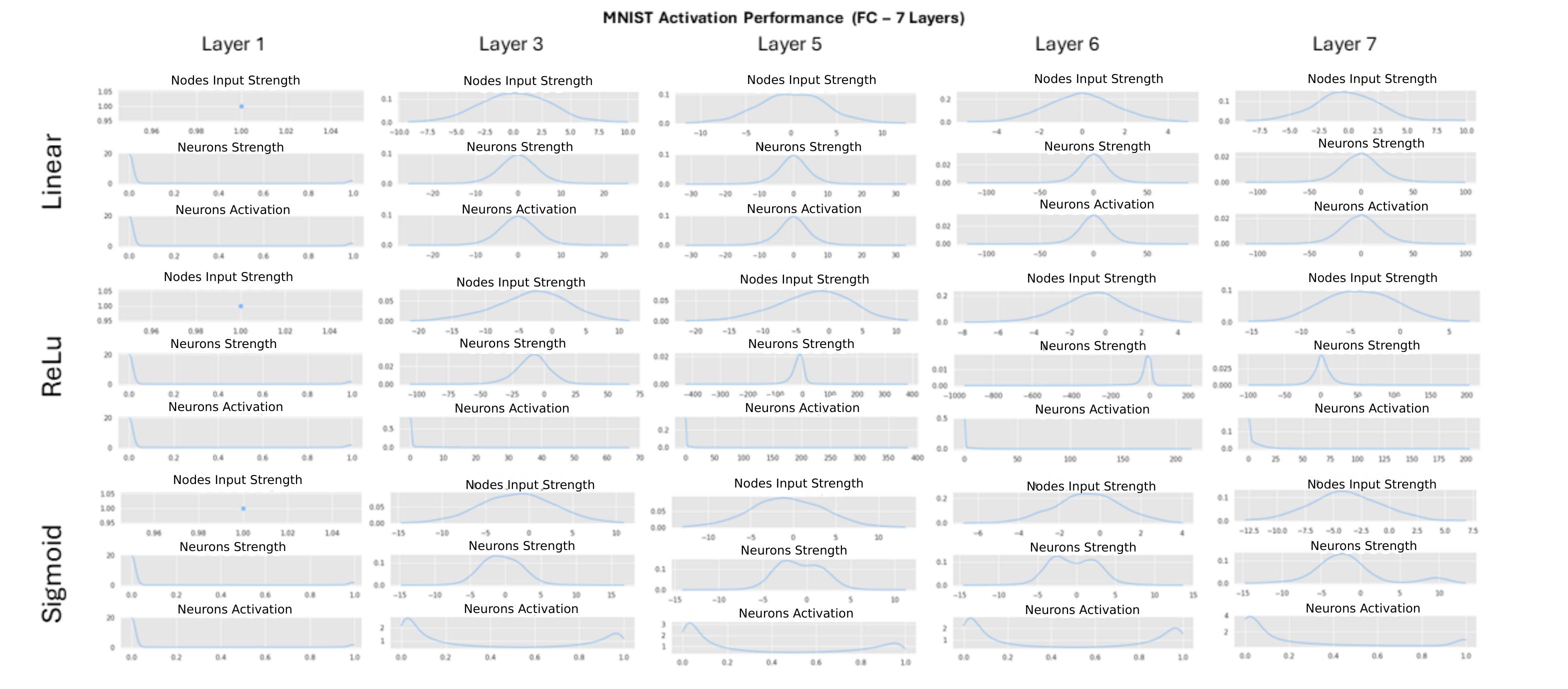}}
    \caption{CNT metrics on MNIST with three activation functions: linear, ReLU and sigmoid (7 hidden layers).}
    \label{fig3}
\end{figure}

\begin{figure}[]
    \centering
    \makebox[\textwidth][c]{\includegraphics[width=1.2\textwidth]{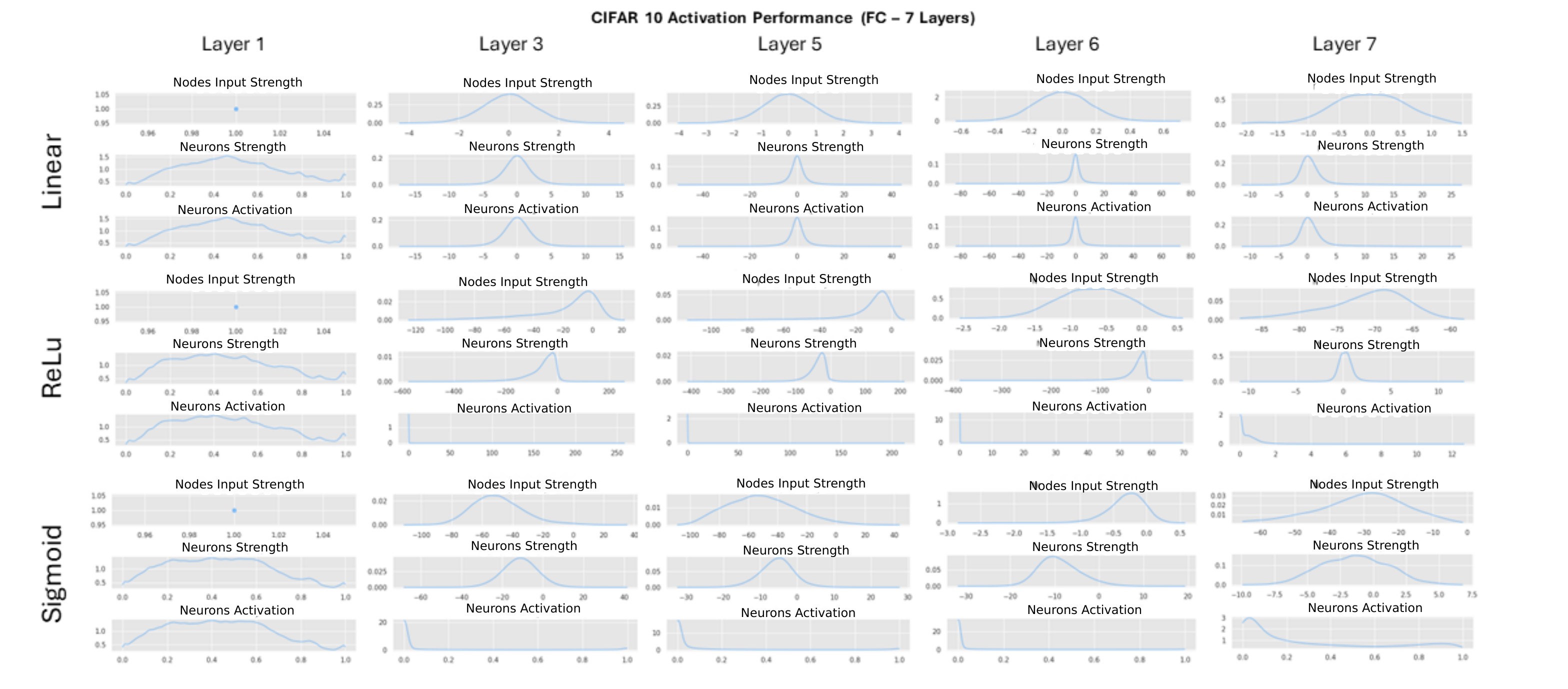}}
    \caption{CNT metrics on CIFAR10 with three activation functions: linear, ReLU and sigmoid (7 hidden layers).}
    \label{fig4}
\end{figure}

\paragraph{\textbf{Networks activation performance}}
Figures \ref{fig3} and \ref{fig4} show how CNT metrics vary as activation functions change when solving MNIST and CIFAR10.
\newline
For MNIST the behaviour of Nodes Strength is very similar for all the activations. Neurons Strength, on the other hand, show the largest support with ReLU activation characterized by long tails. In sigmoid activated layer, bimodality appears in deepest layers.
Neurons Activation is the most discriminative metric, with patterns that characterize uniquely the three activations. In linear networks the distribution is Gaussian, in ReLU it is centered around negative values (which are mapped to $0$), while in sigmoid it presents the typical concentration around the activation's saturation points.
\newline
In CIFAR10, differences between activations and metrics are accentuated. 
Linearly activated Nodes Strength assume a bell shape centred in $0$, while ReLUs are negative skewed. In sigmoids, the support is generally larger than any other activation. 
ReLU and sigmoid both present a monotonic increase in the support of Nodes Strength, followed by a bottleneck at layer $6$. Support returns to precedent magnitude in both the cases.

Neurons Strength for linear activation is centered around $0$ with bell shape, while it is skewed toward negative values for ReLU. Sigmoids are strongly centered around negative values.

Neurons Activation is peaked in $0$ for linear activations, while for sigmoid and ReLU they similarly assume long positive tails and strong concentration on $0$.

\begin{figure}[]
    \centering
    \makebox[\textwidth][c]{\includegraphics[width=1.2\textwidth]{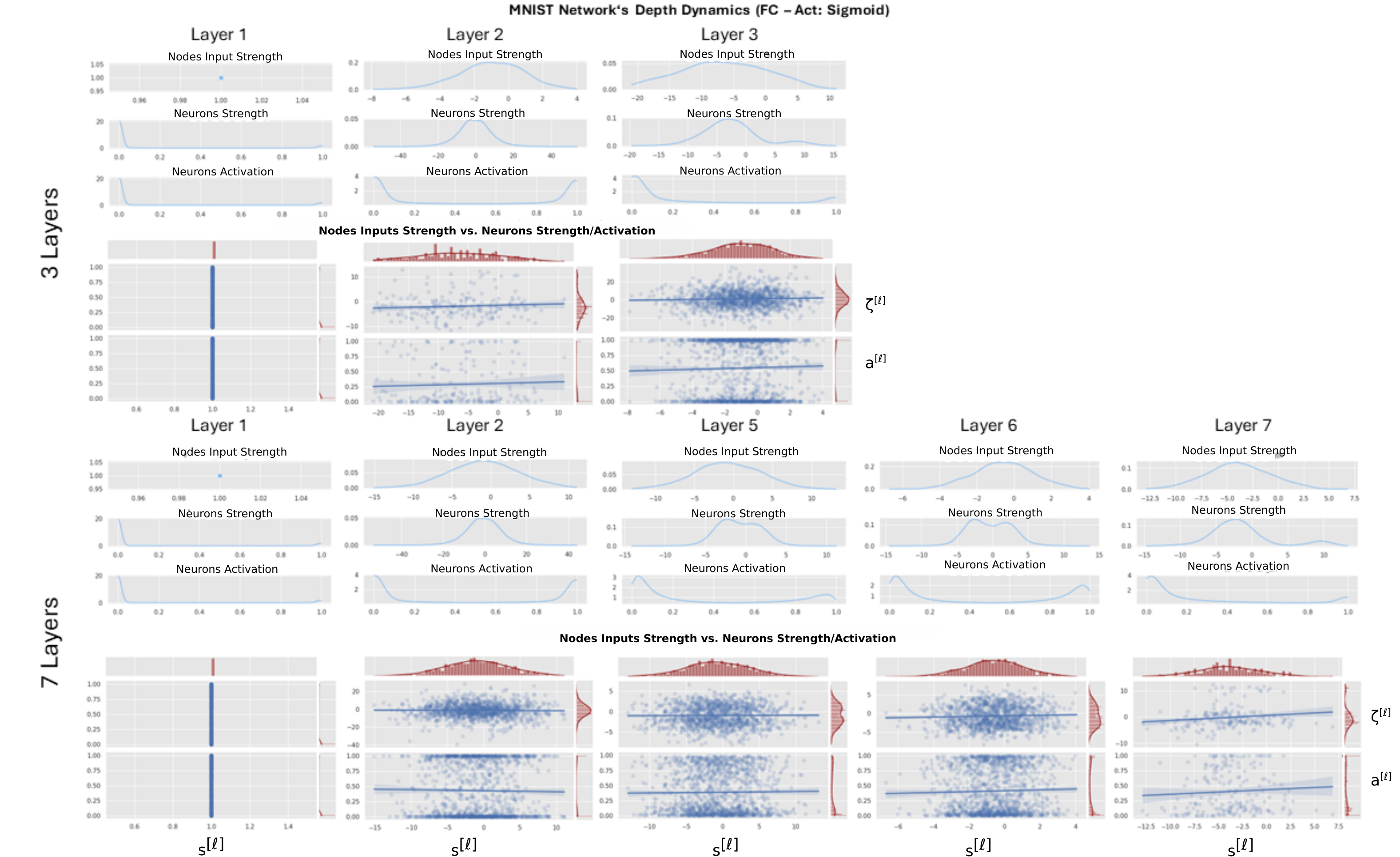}}
    \caption{CNT metrics on MNIST to discriminate shallow (3 layers, top) vs. deep architectures (7 layers, bottom). The activation function is a sigmoid for each layer, excluding the last one which is a softmax.}
    \label{fig5}
\end{figure}

\paragraph{\textbf{Networks' depth}}
Figures \ref{fig4} and \ref{fig5} show CNT metrics sensitivity to the networks' depth, respectively for MNIST and CIFAR10. We report results for DNNs with $3$ (top) and $7$ (bottom) hidden layers.
\newline 
As regards MNIST, we discovered that deep networks have layers whose CNT metrics correspond to those of shallow ones, as in the case of 3 vs. 7-layers networks shown in Figure \ref{fig5}. In fact, the first 3 layers of the shallow networks match with the first, the second and the last layer of the deeper network. We argue that this behavior can be generalized and applicable to different architectures: we reserve to explore this finding in future works as one can study the behavior of complex topologies via an analysis of the building blocks of shallow architectures.
\newline
We found the same observations to be true for CIFAR10, hence we conclude that some learning patterns are invariant between shallow and deep networks.

\begin{figure}[]
    \centering
    \makebox[\textwidth][c]{\includegraphics[width=1.2\textwidth]{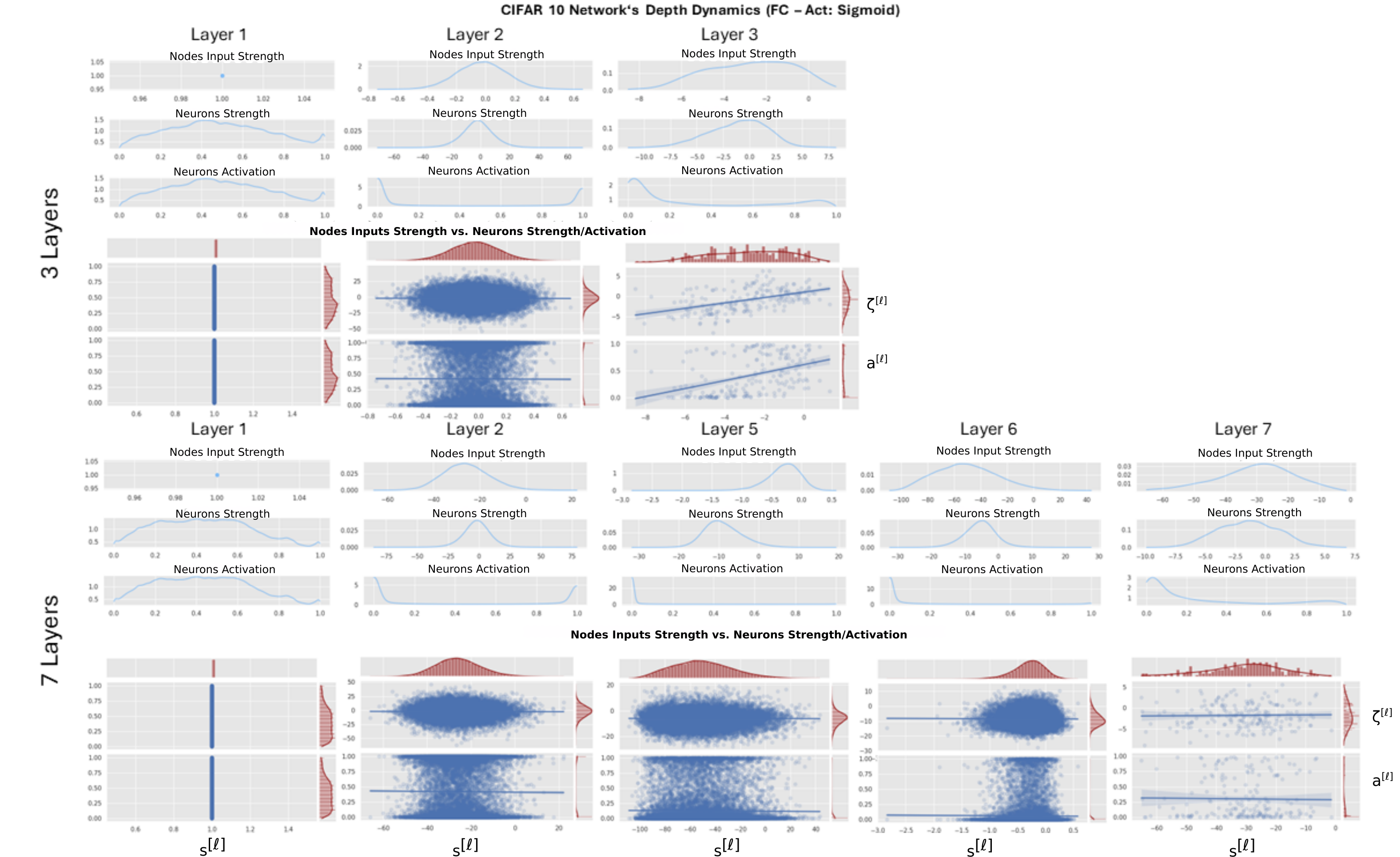}}
    \caption{CNT metrics on CIFAR10 to discriminate shallow (3 layers, top) vs. deep architectures (7 layers, bottom). The activation function is a sigmoid for each layer, excluding the last one which is a softmax.}
    \label{fig6}
\end{figure}

\subsection{Results for Signal Reconstruction}
\begin{figure}[]
    \centering
    \makebox[\textwidth][c]{\includegraphics[width=1.2\textwidth]{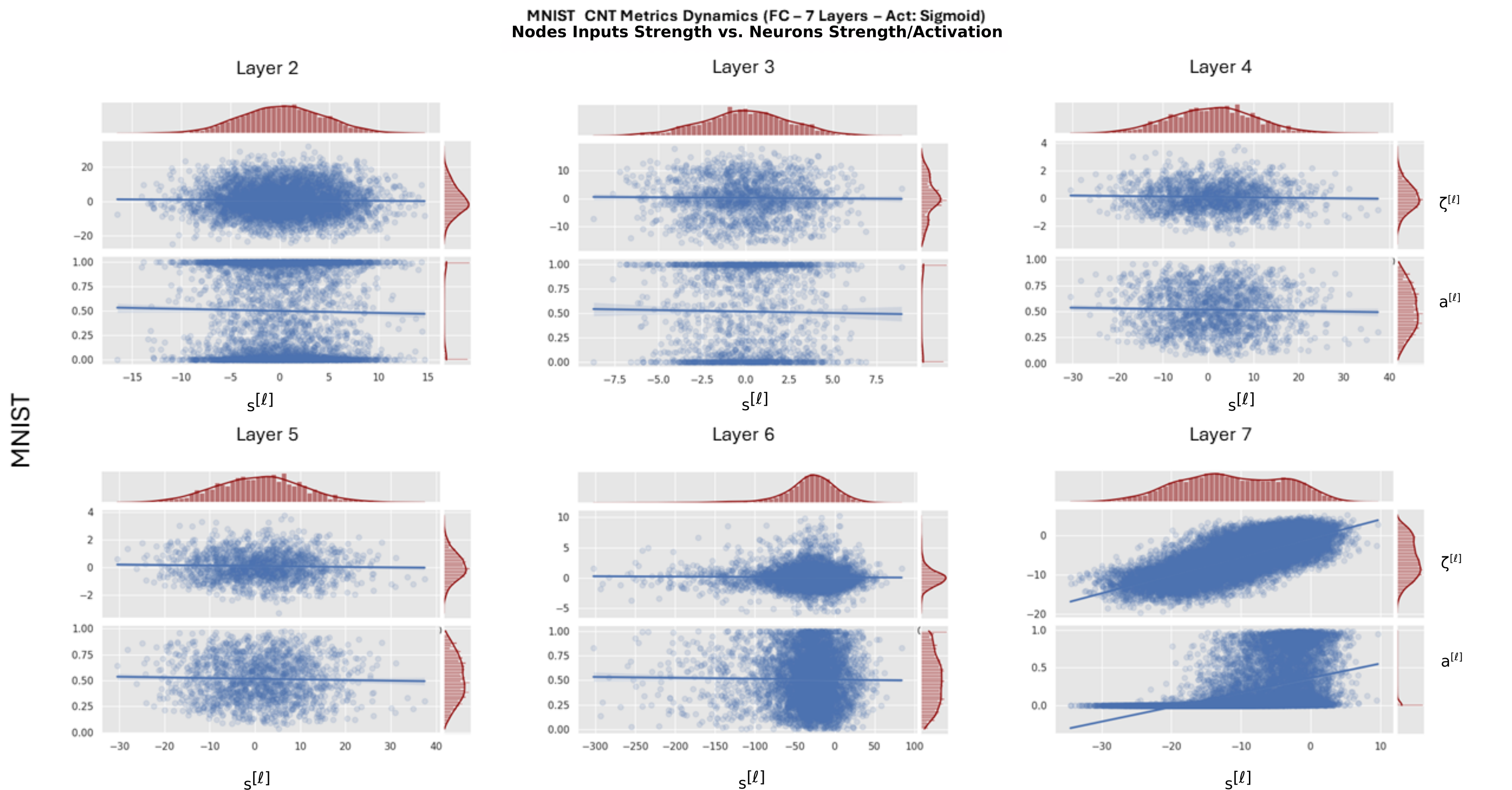}}
    \caption{CNT metrics on MNIST for AutoEncoders. Networks have 7 layers, each activated with a sigmoid, excluding the last one which has a softmax.}
    \label{fig7}
\end{figure}

\begin{figure}[]
    \centering
    \makebox[\textwidth][c]{\includegraphics[width=1.2\textwidth]{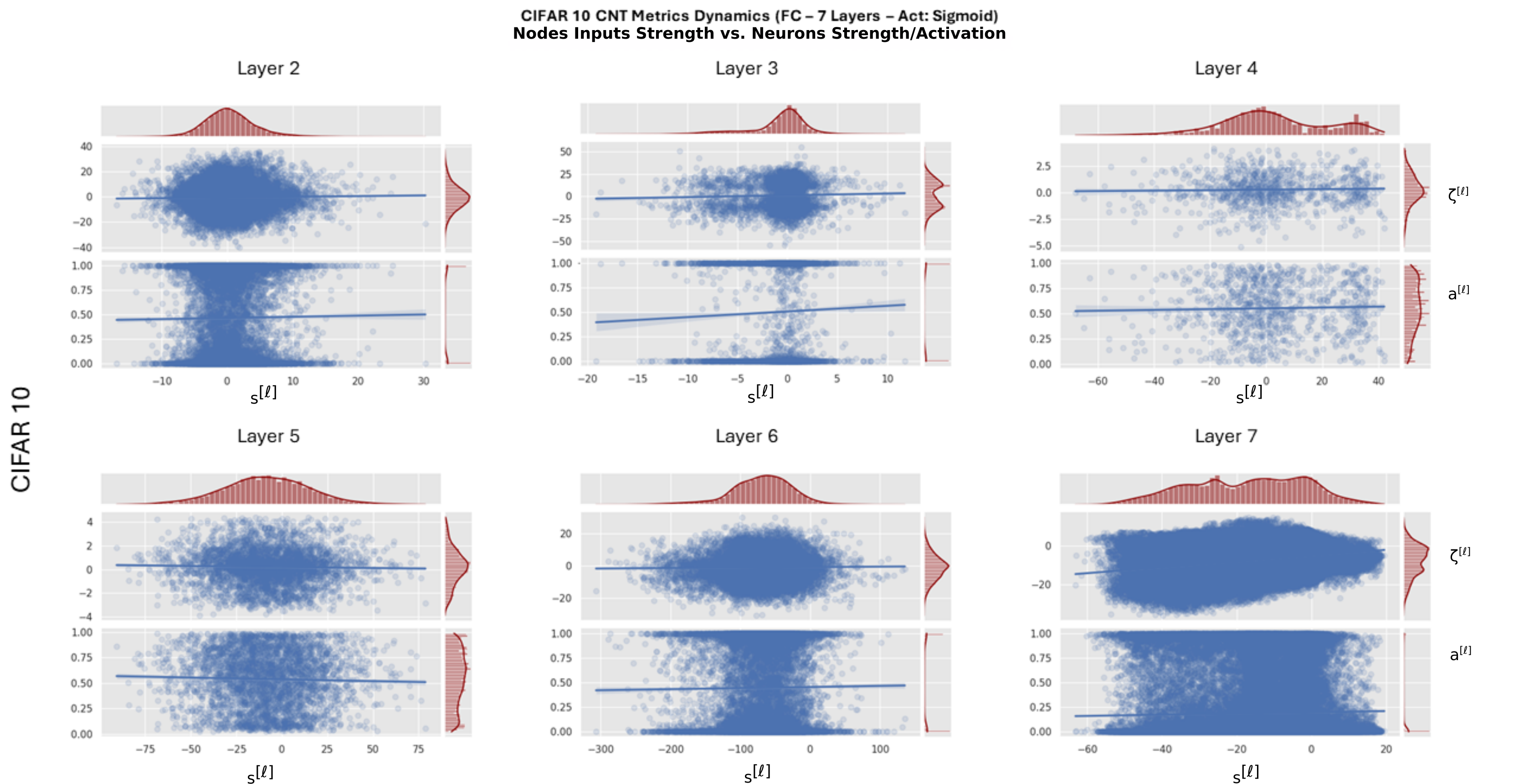}}
    \caption{CNT metrics on CIFAR10 for AutoEncoders. Networks have 7 layers, each activated with a sigmoid, excluding the last one which has a softmax.}
    \label{fig8}
\end{figure}

\paragraph{\textbf{Tasks discrimination. }}
In this paragraph we analyse the CNT metrics for MNIST and CIFAR10 when the network is an Auto-Encoder that takes as input an image and reconstruct it. Cogent results are reported in fig \ref{fig7} and \ref{fig8}. For both the tasks and the networks, the Neurons Activation clearly identifies $3$ distinct phases: a first phase, in layers $1$ and $3$ where the densities are concentrated in $0$ and $1$; another phase where Neurons Activation is uniform (layers $4$, $5$ and $6$ for MNIST, layers $4$ and $5$ for CIFAR10); and a third phase, in the last layers of the networks, where Neurons Activation is concentrated around the saturation values of the sigmoid.

\begin{figure}[]
    \centering
    \makebox[\textwidth][c]{\includegraphics[width=1.2\textwidth]{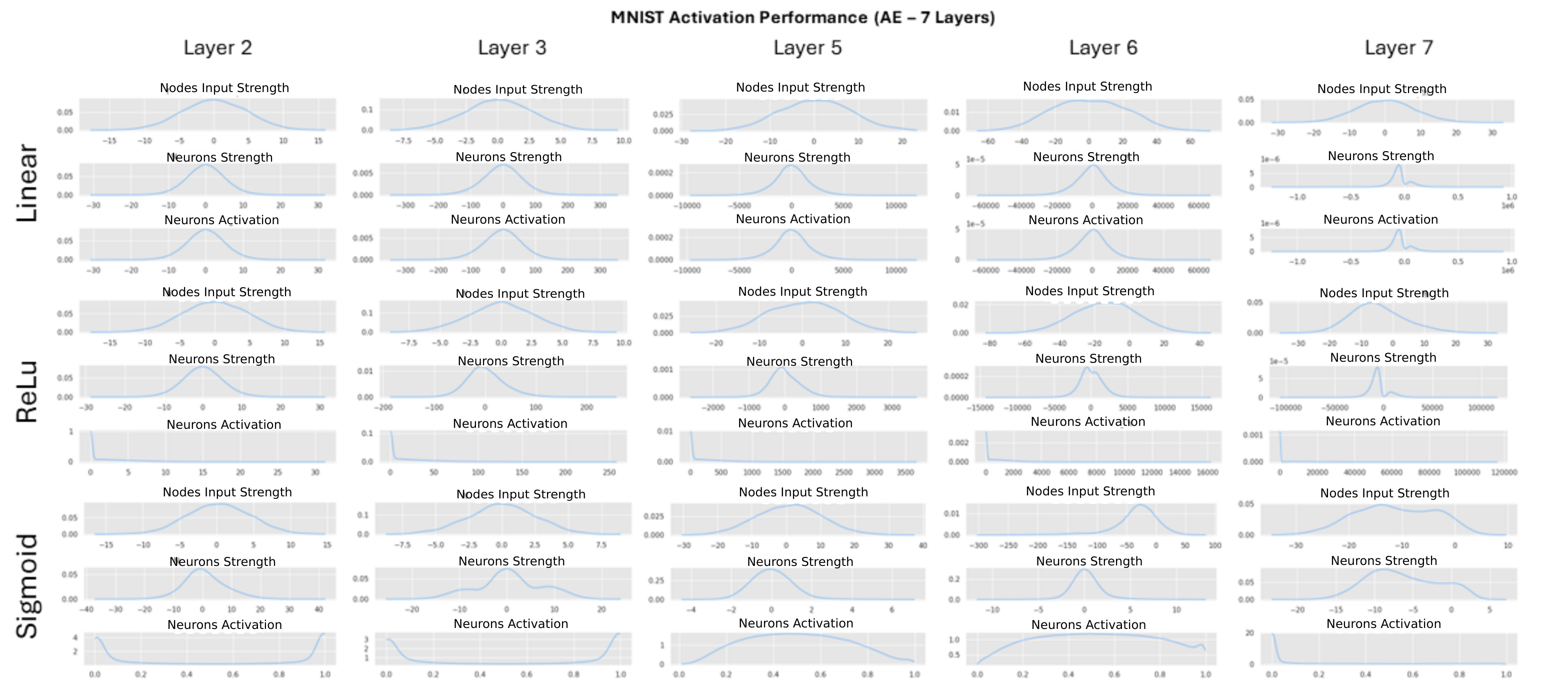}}
    \caption{CNT metrics on MNIST for AutoEncoders with three different activation functions: linear, ReLU and sigmoid (7 hidden layers).}
    \label{fig_9}
\end{figure}

\begin{figure}[]
    \centering
    \makebox[\textwidth][c]{\includegraphics[width=1.2\textwidth]{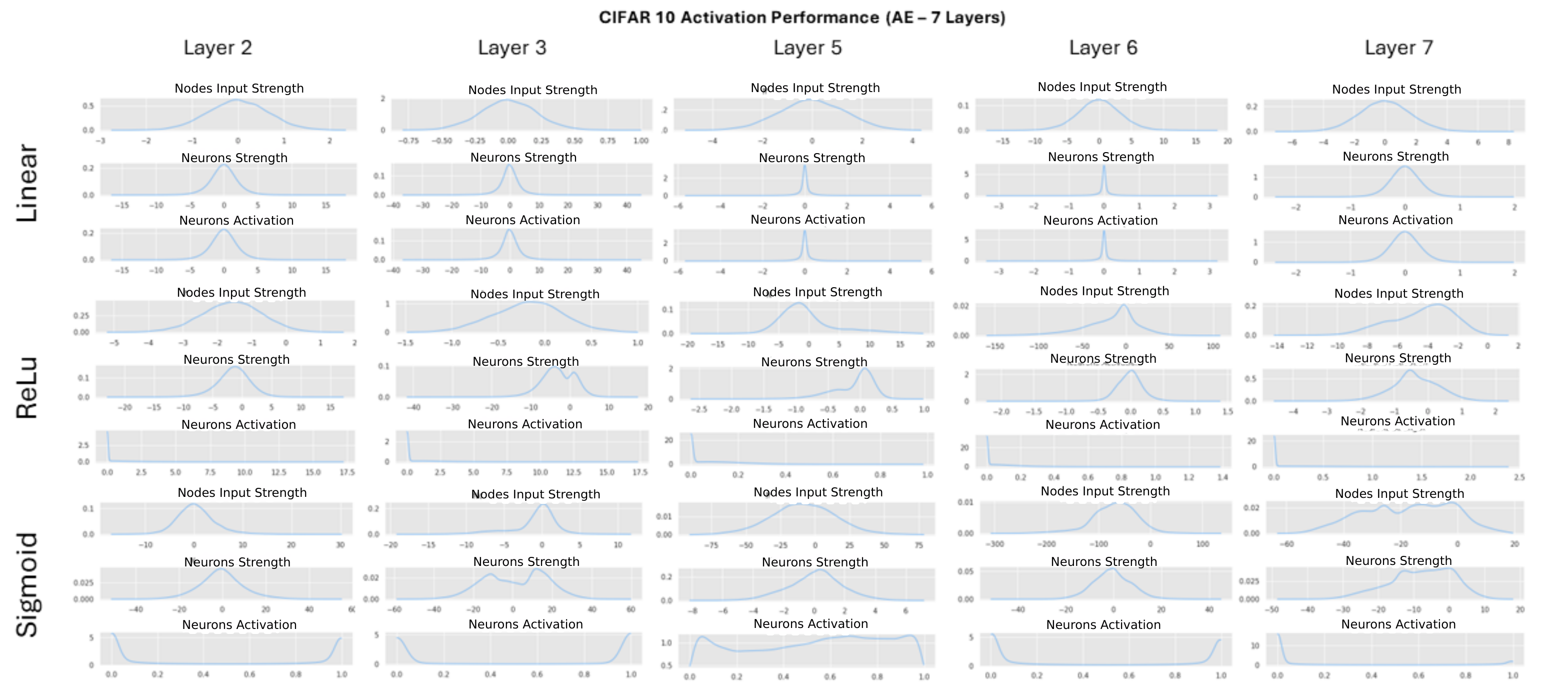}}
    \caption{CNT metrics on CIFAR10 for AutoEncoders with three different activation functions: linear, ReLU and sigmoid (7 hidden layers).}
    \label{fig_10}
\end{figure}

\paragraph{\textbf{Networks activation performance}}  
Fig. \ref{fig_9} and \ref{fig_10} show how CNT metrics vary as activation functions change when solving MNIST and CIFAR10.
\newline 
For MNIST (fig\ref{fig_9}), Neurons Activation show peculiar trends for each of the $3$ activation functions. With linear activation Neurons Activation does not seem to vary from Neurons Strength. In ReLU networks, the distribution concentrates around $0$, while in sigmoids it shows two phases: the former, in layers $2$ and $3$, where the densities concentrate around $0$ and $1$, and the latter, in layers $5$ and $6$, where it becomes uniform. 
\newline 
Neurons Strength exhibits similar distributions for linear and ReLU networks, with large support and thus the presence of extreme values. In sigmoid networks, the support is much concentrated and around 2/3 orders of magnitude less than linear and sigmoid networks.
\newline
In CIFAR10 (Fig.\ref{fig_10}), linear and ReLU activated networks show a reduced support of Neurons Strength metrics, compared to sigmoid, against the trend of MNIST networks described previously. Finally, the phases of the Neurons Strength are peculiar to each activation activation function and uniquely characterize each network. 

\begin{figure}[]
    \centering
    \makebox[\textwidth][c]{\includegraphics[width=1.2\textwidth]{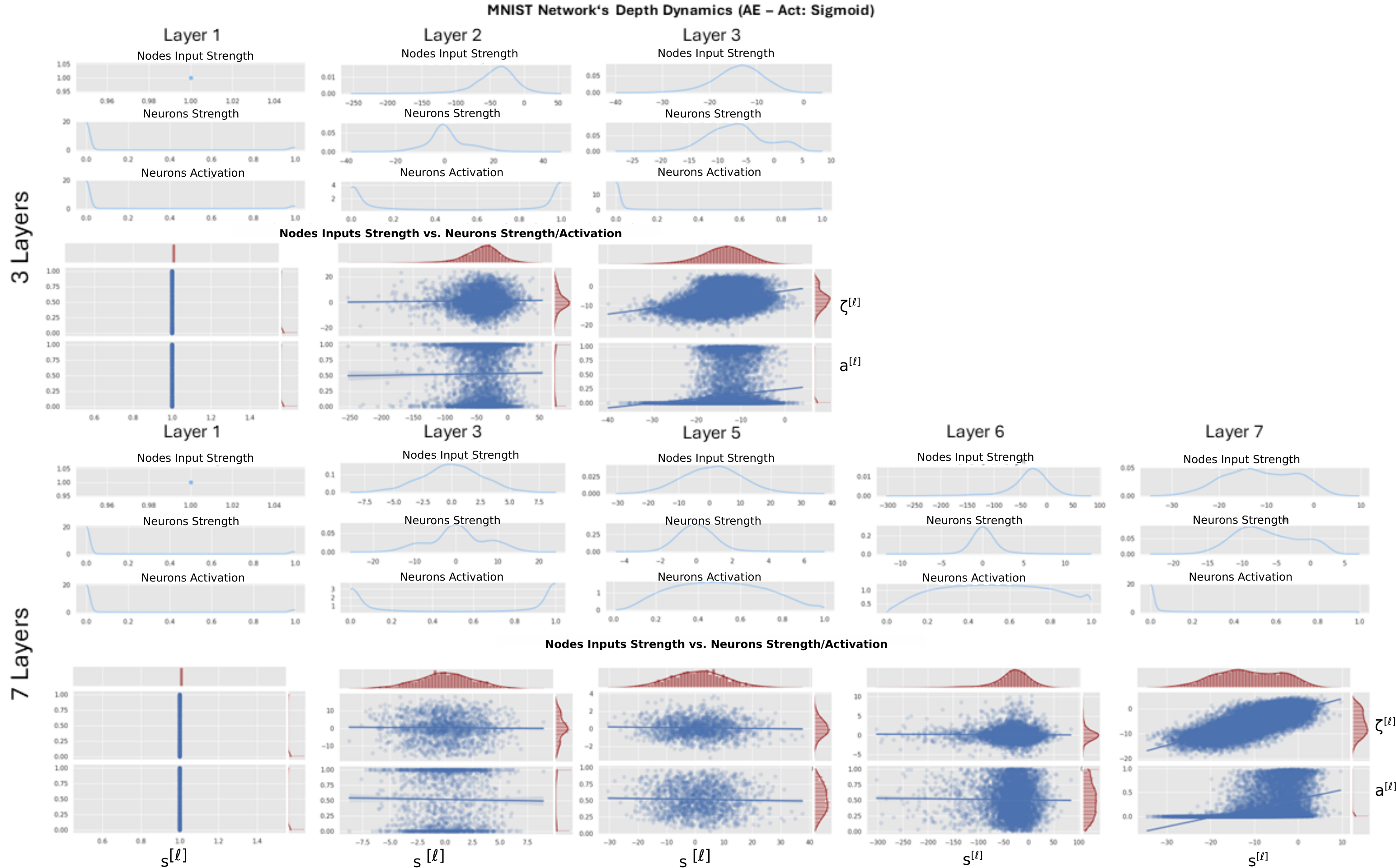}}
    \caption{CNT metrics on MNIST for AutoEncoders to discriminate shallow (3 layers, top) vs. deep architectures (7 layers, bottom). The activation function is a sigmoid for each layer, excluding the last one which is a softmax.}
    \label{fig11}
\end{figure}

\begin{figure}[]
    \centering
    \makebox[\textwidth][c]{\includegraphics[width=1.2\textwidth]{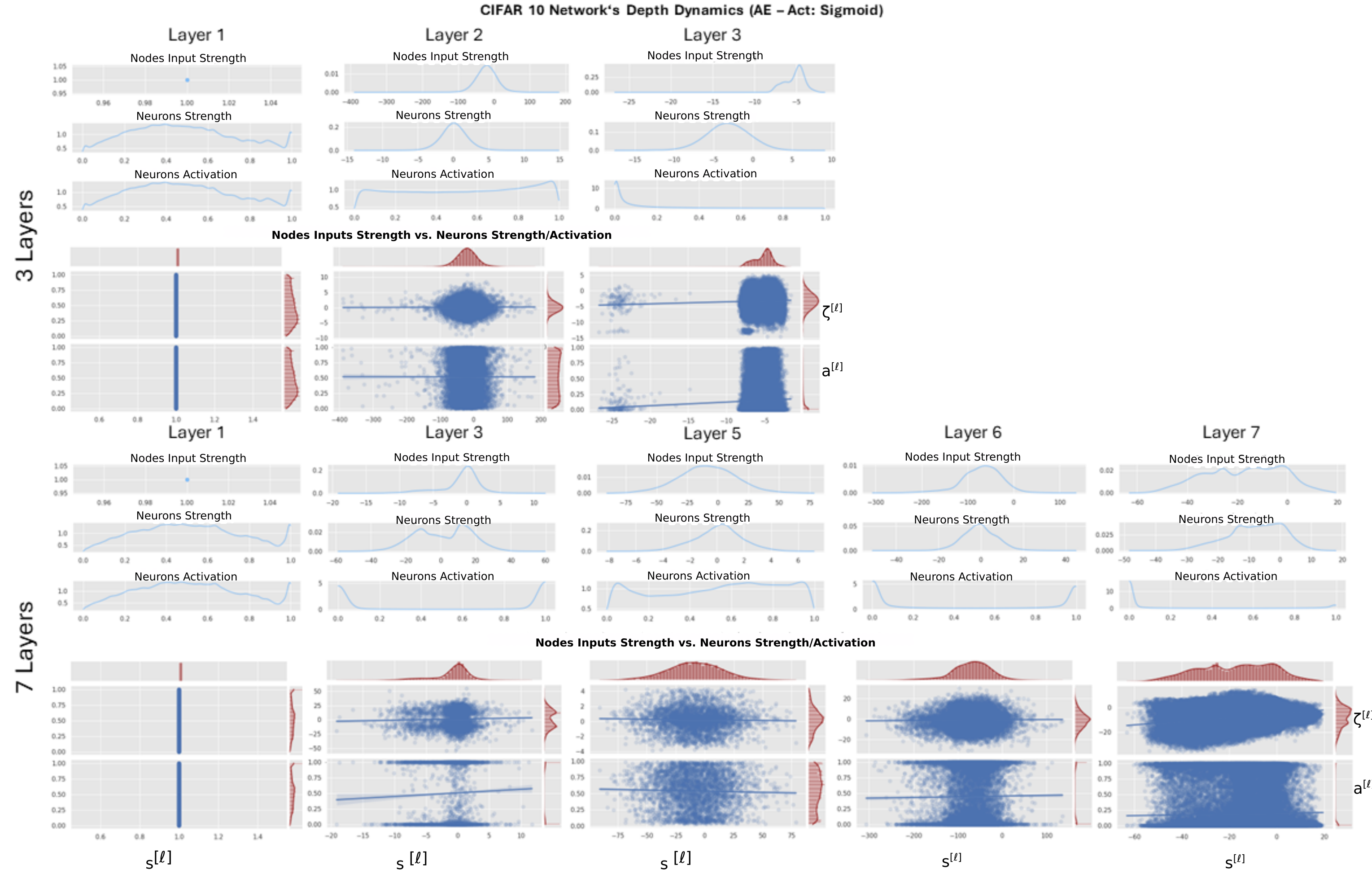}}
    \caption{CNT metrics on CIFAR10 for AutoEncoders to discriminate shallow (3 layers, top) vs. deep architectures (7 layers, bottom). The activation function is a sigmoid for each layer, excluding the last one which is a softmax.}
    \label{fig12}
\end{figure}

\paragraph{\textbf{Networks depth}}  
Similarly to what we reported previously, metrics in shallow networks have a correspondent in deeper architectures. As reported in Fig.\ref{fig11} (MNIST), Neurons Strength and Activation of $3$-layers networks behave similarly to layers $3$ and $7$ of the deeper network.
\newline
As regards CIFAR10, whose results we report in \ref{fig12}, similarities between shallow and deep networks are less accentuated but still present.

\subsection{High and Low performing FC networks}
\begin{figure}[]
    \centering
    \makebox[\textwidth][c]{\includegraphics[width=1.2\textwidth]{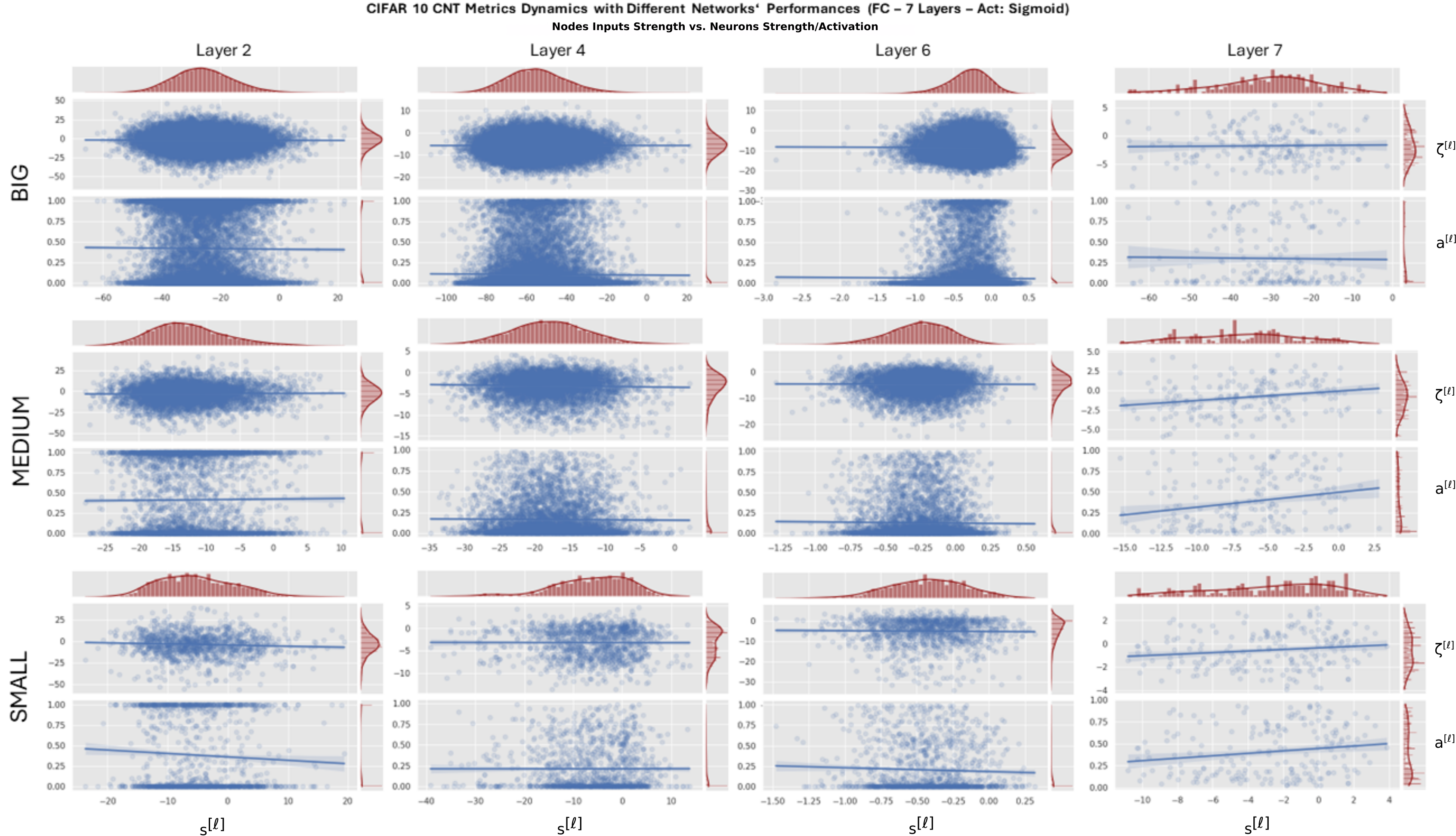}}
    \caption{CNT metrics for FC networks with different performances on CIFAR10.}
    \label{fig13}
\end{figure}

Table \ref{tab:cifar10fc} we report details of FC networks of different size, trained on CIFAR10.\footnote{As already mentioned, this Section encompasses experiments with networks trained on CIFAR10 as obtaining low-performing networks on MNIST without introducing artificially distortive bottlenecks is hard, as empirically evidenced in previous works \cite{LAMALFA2021}.}.
In Fig.\ref{fig13} is shown a comparison of the CNT metrics of such networks to evidence possibly diverging behaviors between big, medium and small architectures. 
\newline 
The CNT metrics, despite differences in the distributions between big, medium and small architectures, share similar trends: a bottleneck in Nodes Input Strength is present in layer $6$ where the support of the Strength is significantly reduced, then it returns to a range of values comparable to the precedent layers.
Neurons Strength are bell-shaped  for all the layers excluded the last one, which flattens along its support. Neurons Activation are mainly distributed around $0$, with the remaining minority of values centered in $1$.

\begin{table}[]
\centering
\begin{tabular}{|l|l|l|l|l|}
\hline
\textbf{Architecture} & \textbf{\# Layers} & \textbf{Activation} & \textbf{\# Parameters} & \textbf{Accuracy} \\ \hline
\textbf{Small}  & 7 & sigmoid & 28K  & 0.38 \\ \hline
\textbf{Medium} & 7 & sigmoid & 450K & 0.49 \\ \hline
\textbf{Big}    & 7 & sigmoid & 7M   & 0.54 \\ \hline
\end{tabular}
\caption{The Table reports details about the FC architectures used to solve CIFAR10. Experiments on 3 different architectures are reported, namely big, medium and small.
While one can argue that accuracy is low for CIFAR10 (even for big networks), we note that we trained vanilla FC architectures that are compared to spot \textbackslash{}textit\{accuracy-gaps\} revealed by the CNT metrics. We point out that we couldn't find (in our or other experimental evaluations in research literature) any vanilla FC with accuracy greater than 0.6 \cite{LIN2015}.}
\label{tab:cifar10fc}
\end{table}

\section{Conclusions}
This paper presents a new framework for interpreting neural networks via Complex Network Theory.
\newline 
We introduce a formal framework that turns a neural network into an equivalent Complex Network that can be studied at a weights, neurons and layers level.
\newline 
We perform an extensive experimental evaluation on populations of networks: we find that our framework discriminates models: (i) initialized with different architectures (FC and AE); (ii) Trained on different tasks (MNIST and CIFAR10) and objectives (pattern recognition vs. signal reconstruction); (ii) where the depth and the activation function vary. We further show how CNT metrics represent performing vs. non-performing networks. 
\newline 
In future works, we will extend this framework to advanced architectures (e.g., attention, residual-networks, etc.), both theoretically and empirically.

\newpage
\appendix

 \bibliographystyle{elsarticle-num} 
 \bibliography{bibliography_CNT}





\end{document}